\documentclass[sigconf]{acmart}
\AtBeginDocument{%
  }
\usepackage{graphicx}
\usepackage{float}
\usepackage{amsmath}
\usepackage{color}
\usepackage{multicol}
\usepackage{multirow}
\usepackage{mathtools}
\usepackage{hyperref}
\usepackage{svg} % 需使用包
\usepackage[ruled,linesnumbered]{algorithm2e}
\copyrightyear{2024}
\acmYear{2024}
\setcopyright{acmlicensed}
\acmConference[MMASIA '24]{ACM Multimedia Asia}{December 3--6, 2024}{Auckland, New Zealand}
\acmBooktitle{ACM Multimedia Asia (MMASIA '24), December 3--6, 2024, Auckland, New Zealand}
\acmDOI{10.1145/3696409.3700201} \acmISBN{979-8-4007-1273-9/24/12}

\begin{document}

\title{DiffuseST: Unleashing the Capability of the Diffusion Model for Style Transfer}

\author{Ying Hu}
\authornote{Equal contribution.}
\email{ying.hu@nuaa.edu.cn}
\affiliation{
  \institution{Nanjing University of Aeronautics and Astronautics}
  \city{Nanjing}
  \country{China}
}

\author{Chenyi Zhuang}
\authornotemark[1]
\email{chenyi.zhuang@nuaa.edu.cn}
\affiliation{
  \institution{Nanjing University of Aeronautics and Astronautics}
  \city{Nanjing}
  \country{China}
}

\author{Pan Gao}
\authornote{Corresponding author.}
\email{pan.gao@nuaa.edu.cn}
\affiliation{
  \institution{Nanjing University of Aeronautics and Astronautics}
  \city{Nanjing}
  \country{China}
}


\begin{abstract}
  Style transfer aims to fuse the artistic representation of a style image with the structural information of a content image. Existing methods train specific networks or utilize pre-trained models to learn content and style features. However, they rely solely on textual or spatial representations that are inadequate to achieve the balance between content and style. In this work, we propose a novel and training-free approach for style transfer, combining textual embedding with spatial features and separating the injection of content or style. Specifically, we adopt the BLIP-2 encoder to extract the textual representation of the style image. We utilize the DDIM inversion technique to extract intermediate embeddings in content and style branches as spatial features. Finally, we harness the step-by-step property of diffusion models by separating the injection of content and style in the target branch, which improves the balance between content preservation and style fusion. Various experiments have demonstrated the effectiveness and robustness of our proposed DiffeseST for achieving balanced and controllable style transfer results, as well as the potential to extend to other tasks. Code is available at \href{https://github.com/I2-Multimedia-Lab/DiffuseST}{https://github.com/I2-Multimedia-Lab/DiffuseST}.
\end{abstract}






\maketitle
\begin{figure}
\centering
\includegraphics[width=\linewidth]{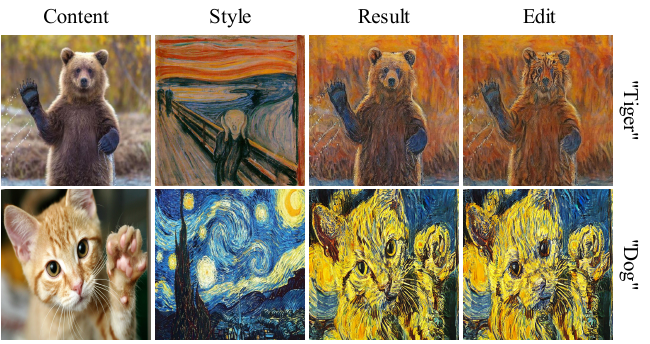}
\vspace{-20pt}
\caption{Style transfer (column 3) and extended image editing (column 4) results by our method.}
\vspace{-18pt}
\label{banner}
\end{figure}
\section{Introduction}
Every image has its specific content and unique style. When we are interested in a particular style and want to turn our own photos into new images combined with that style, it is called style transfer: a technique to inject the style from one image into the content of another image.
The main goal of style transfer is to inject the artistic characteristics of the style image into the content image while not destroying the essential structure of the content, which is not a simple task and has been widely studied. 

Starting from Gatys' pioneering work \cite{gatys2016image} that first introduced Convolutional Neural Networks (CNNs) to style transfer, numerous works \cite{chen2016fast,gu2018arbitrary,huang2017real} have emerged to improve the quality of stylization. The style characteristics are extracted by delicately designed networks like CNNs, attention modules, and Transformers. Nevertheless, it is still challenging to balance content preservation and style injection. Existing methods fail to produce balanced stylized images, showing (1) excessive artistic features to deviate from the original content, or (2) firm content features and inadequate style features. Meanwhile, target artistic styles are often quite subtle, leading to difficulty for trained networks to grasp them well. These limitations have significantly affected stylization quality. 

Typically, text-to-image diffusion models \cite{rombach2022high} with the ability to interact text with image features have provided multi-modal power for various vision tasks \cite{hertz2022prompt, brooks2023instructpix2pix, ruiz2023dreambooth}.
Pre-trained diffusion models have been explored as learning the text embedding of the style image \cite{zhang2023inversion}, or designing objective functions to fine-tune the denoising U-Net \cite{yang2023zero}. However, the practicality of these methods is limited by the high cost of time and memory required for optimization.
Furthermore, we notice that the step-by-step denoising property of diffusion models is still underused for the style transfer task, which has been extensively investigated in image editing \cite{meng2021sdedit, wang2023not, xu2023inversion} to balance fidelity and editability. We then come up with the question: \textbf{Can we leverage the step-by-step ability of pre-trained diffusion models to achieve a balanced stylization by separating the injection of content and style features}?

In this work, we propose DiffuseST for style transfer, exploiting the step-by-step capability of the pre-trained diffusion model. Inspired by previous works, we combine spatial and textual features to balance content details and artistic styles. For textual representation, we adopt the multimodal BLIP-2 encoder from BLIP-Diffusion  \cite{li2024blip} to produce the text-aligned semantic feature of the style image. For spatial representation, we follow \cite{tumanyan2023plug} that uses the DDIM inversion \cite{ho2020denoising} technique to extract the spatial features of the content and style images in the content and style branches, respectively. Benefiting from two expressive representations, DiffuseST achieves efficient and high-quality stylization in a training-free manner. Our main contributions can be summarized as follows:

\begin{itemize}
\item  We propose a novel and training-free approach for style transfer that combines the textual and spatial representations of the style and content images. 
% A training-free approach is proposed that enables style transfer without fine-tuning the training diffusion.
\item  We harness the step-by-step denoising ability of pre-trained diffusion models by separating the injection of content and style representation in the target branch.
% We encode the style image as a sequence, and successfully implement style injection. And we use the content injection module and style injection module to achieve a balance between content retention and style injection.
\item Extensive experiments are conducted for comparison and evaluation, demonstrating the superiority and effectiveness of the proposed method in balanced synthesis.
\end{itemize}
\section{Related Work}
$\textbf{Image Style Transfer}$ has long been a hot research topic in image generation. Traditional methods such as Gatys' work \cite{gatys2016image} train specific deep learning networks, which can be time-consuming. Subsequently, fast style transfer methods align content and style features by matching the statistics of mean, variance, and covariance, exemplified by WCT \cite{li2017universal} and AdaIN \cite{huang2017arbitrary}. Other works \cite{zhu2017unpaired, isola2017image} adopt Generative Adversarial Networks (GANs) with the discriminator to ensure synthesis consistency. The later emerged attention mechanisms \cite{liu2021adaattn,yao2019attention} and Transformer \cite{deng2022stytr2} networks provide new insights into style transfer, and have achieved significant improvements. Recently, multi-modal networks like CLIP \cite{radford2021learning} have inspired researchers to utilize text information as additional conditions for style transfer \cite{kwon2022clipstyler}. 
However, the expression of natural language is limited, restricting the range of available styles and failing to comprehend images with obscure artistic styles.
\\
$\textbf{Diffusion Models}$ have facilitated various vision tasks, such as image super-resolution \cite{wang2023exploiting,lin2023diffbir,yang2023pixel} and image editing \cite{couairon2022diffedit,kawar2023imagic,avrahami2022blended}, as well as style transfer. InST \cite{zhang2023inversion} turns style images into text embedding as the conditional input to the diffusion model. VCT \cite{cheng2023general} seeks textual embeddings of both content and style images. StyleDiffusion \cite{wang2023stylediffusion} explicitly extracts content information and implicitly learns complementary style information for interpretable and controllable content-style decoupling and style transfer. Yet, these methods are far from optimal as detailed content and complicated style features cannot be captured solely by text embeddings. Moreover, the existing diffusion-based methods require several steps for fine-tuning or optimizing, which are time-consuming. In contrast, our proposed method is training-free, balancing textual and spatial content and style information, which can be more efficient and effective. 

\begin{figure*}[htbp]
\centering
\includegraphics[width=\linewidth]{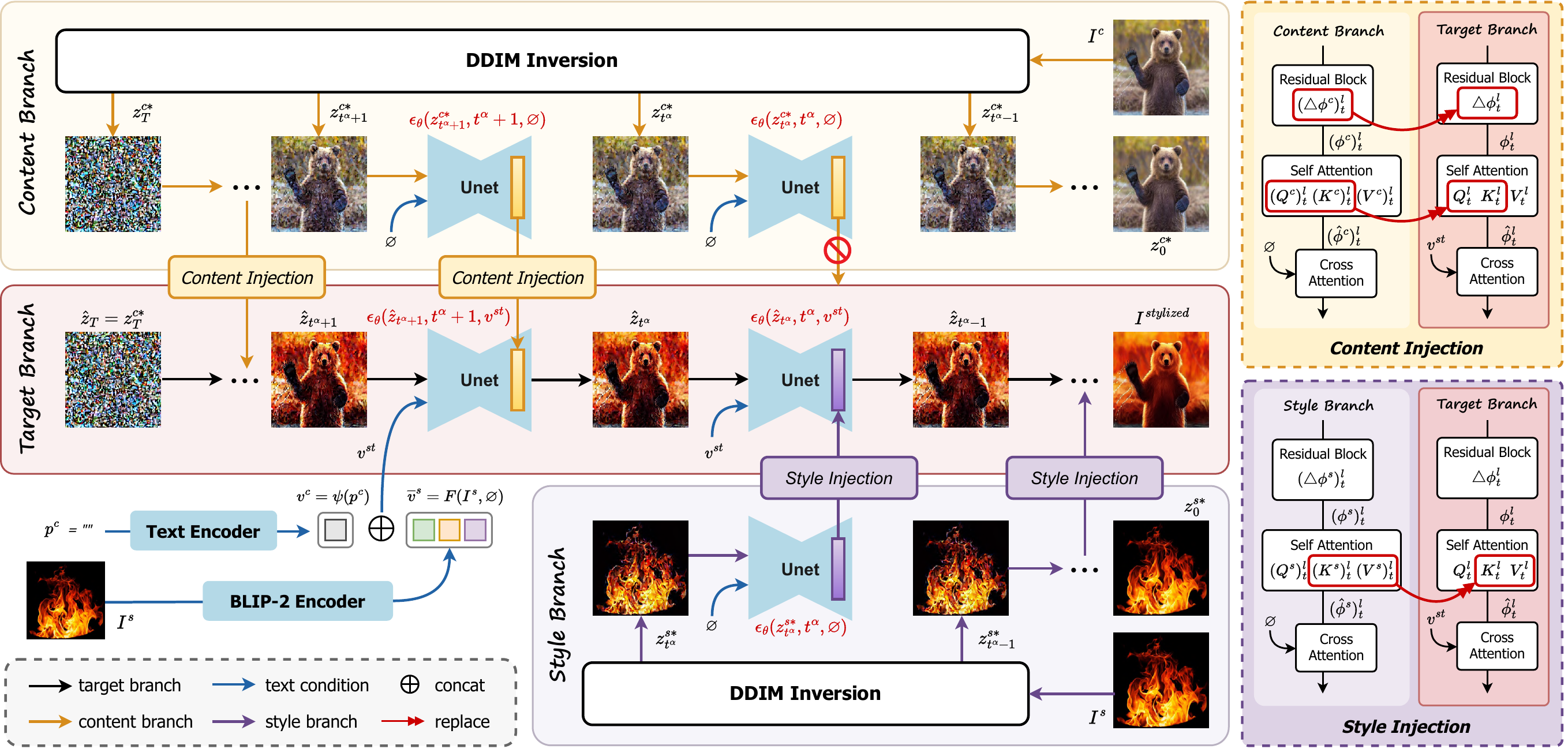}
\caption{Overall framework of DiffuseST. The target branch is to perform style transfer guided by textual and spatial representations of two images. We adopt the BLIP-2 encoder to produce text-aligned features of the style image. We utilize the DDIM inversion technique and extract inner spatial features in the content and style branches, respectively. The content and style spatial injections are separated at different steps in the target branch to achieve balanced stylization.}
\label{network}
\end{figure*}
\section{Method}
Previous methods that solely rely on textual or spatial features fail to strike the stylization balance. In this work, we are inspired to combine two representations of the content image $I^c$ and the style image $I^s$. Specifically, we adopt the BLIP-2 encoder from BLIP-Diffusion \cite{li2023blip} to produce text-aligned embeddings of $I^s$ (Section \ref{textual_representation_extraction}). To extract spatial features, we apply DDIM inversion and preserve U-Net's inner features in the content and style branches, respectively (Section \ref{spatial_representation_extraction}). 
Furthermore, we take full advantage of diffusion models' step-by-step ability by separating the injection of content and style in the target branch (Section \ref{unified}). Figure \ref{network} illustrates the overall framework. Recall that our method is training-free and all used models are frozen during generation. 

\subsection{Preliminary}

Our method is built upon the latent diffusion model \cite{rombach2022high}. It performs forward and backward processes in the latent space instead of the image space. Specifically, the forward process adds Gaussian noise to the encoded image, i.e., latent, and the reverse process iteratively removes the noise from the latent to generate a new image. Appendix ~\ref{sec:Preliminary} provides a detailed background of diffusion models. To modify real images, we adopt the DDIM inversion \cite{song2020denoising, dhariwal2021diffusion} that reverses the ODE process in small steps:
\begin{equation}
\begin{aligned}
\label{inversion}
z^*_{t+1} = \sqrt{\frac{\alpha_{t+1}}{\alpha_t}}z^*_t+(\sqrt{\frac{1}{\alpha_{t+1}}-1}-\sqrt{\frac{1}{\alpha_t}-1})\cdot\epsilon_\theta(z^*_t,t,v)
\end{aligned}
\end{equation}

For simplicity, we denote the sequence of latent variables from the forward process as $\{z_t\}^T_{t=1}$, from the reverse process as $\{\hat{z}_t\}^T_{t=1}$, and from DDIM inversion as $\{z^*_t\}^T_{t=1}$.
% \begin{equation}
% \begin{aligned}
% z_t = \sqrt{\alpha_t}z_0+\sqrt{1-\alpha_t}\epsilon, \epsilon \sim \mathcal{N}(0,1)
% \end{aligned}
% \end{equation}
% where $z_0$ is the latent, $\epsilon$ is the added Gaussian noise, and $\alpha_t:= \prod_{i=1}^{t}(1-\beta_i)$ is the hyper-parameter determined by the schedule $\beta_0,...,\beta_T\in (0,1)$. 

% \textbf{Text-to-Image Diffusion Models}
% In this work, we are particularly interested in pre-trained Latent Diffusion Models (LDMs) conditioned on text prompts. Instead of the image $x_0$ sampled from the real data distribution, the source latent $z_0=\mathcal{E}(x_0)$ is encoded by the encoder $\mathcal{E}$. The image can be generated by the decoder as $\hat{x}_0=\mathcal{D}(\hat{z}_0)$, where $\hat{z}_0$ is the denoised latent by a trained diffusion model $\epsilon_\theta$. Each diffusion step $t$ is conditioned on the text prompt $\mathcal{P}$ as:
% \begin{equation}
%     \hat{z}_{t-1}=\epsilon_\theta(\hat{z}_t,t,v)
% \end{equation}
% where $v=\psi(\mathcal{P})$ is the conditional text embedding encoded by the pre-trained CLIP \cite{radford2021learning} text encoder $\psi$.

\subsection{Textual Representation Extraction}
\label{textual_representation_extraction}
Advanced text-to-image diffusion models generate images conditioned on natural language. However, artistic styles can be difficult to describe in text. Yet, most pre-trained models (e.g., Stable Diffusion) do not allow image features as the conditional input. 

In this work, we consider BLIP-Diffusion \cite{li2024blip}, a novel network for subject-driven editing, which adopts BLIP-2 \cite{li2023blip} encoder to produce text-aligned visual features. The encoded features contain rich generic information and can express the semantic characteristics of style images. For simplicity, we denote the pre-trained BLIP-2 encoder as $\mathcal{F}$, to transform visual features of the style image to text-aligned representation. In addition, we keep the vanilla text embedding produced by CLIP \cite{radford2021learning} text encoder $\psi$ in the pre-trained diffusion model to enhance the content information. The above two text embeddings are concatenated as:
\begin{equation}
\label{st_emb}
    v^{st}=\mathrm{Concat}(v^c,\bar{v}^s)\quad
\mathrm{where}\quad
    v^c=\psi(\mathcal{P}^c), \bar{v}^s=\mathcal{F}(I^s,``")
\end{equation}
where $\mathcal{P}^c$ is an optional guided text with respect to the structural information of the content image $I^c$. In practice, we set the text prompt $\mathcal{P}^c$ to the null text $``"$ to avoid manual labeling, while it can be manually specified to extend to editing task (see Figure \ref{text_edit}). Notice that we also use the null text $``"$ as the input of the BLIP encoder $\mathcal{F}$ to capture global and abstract style features.
% The alternative text $\mathcal{P}^{edit}$ indicates that the manipulation by our approach preserves the editing ability in text-to-image diffusion models, verifying the stability and robustness of our method.

% \begin{figure}[t]
% \centering
% \includegraphics[width=0.92\linewidth]{assets/attention.pdf}
% \vspace{-10pt}
% \caption{The self-attention module in (a) the original U-Net, (b) the revised version with content injection, or (c) with style injection. Note that the replaced elements in content injection are different from those in style injection.}
% \label{attention}
% \end{figure}

\subsection{Spatial Representation Extraction}
\label{spatial_representation_extraction}
Given a clean image, the diffusion forward process progressively adds Gaussian noise that produces intermediate latent $\{z_{t}\}^{T}_{t=1}$. The diffusion U-Net $\epsilon_\theta$ can remove noise from the latent step-by-step as $\hat{z}_t=\epsilon_\theta(\hat{z}_t,t,v)$, where $v$ is the conditional text embedding. However, directly applying the forward and reverse diffusion processes will cause inconsistencies between the input and the generated images. To maintain the structure of $I^c$, we adopt the DDIM inversion technique to achieve better content preservation. 

Intuitively, with the text-aligned embedding $v^{st}$ of the style image, text-guided reverse diffusion can be performed on the noised content image $z^{c*}_T$ obtained by DDIM inversion. However, the generated image will inevitably introduce undesired changes in the content structure (see Figure \ref{style injection}). We suppose that the extracted representation $\bar{v}^s$ involves redundant information, and the reverse process without additional constraint from the content image can lose some essential content details. Therefore, we are motivated to deep into the diffusion architecture to extract the spatial features and better guide the stylization. Speciﬁcally, each layer of the pre-trained diffusion U-Net consists of a residual block, a self-attention module to enhance representation, and a cross-attention module to interact with text condition. In this work, we particularly focus on the former two blocks, which have been well-investigated for containing semantic spatial layout and structure of the generated image \cite{cao2023masactrl, tumanyan2023plug}. For layer $l$ at step $t$, the residual block with the intermediate feature $\Delta \phi^{l}_t$ will output $\phi^{l}_t$. Then a self-attention module enhances the image representation:
\begin{equation}
\begin{aligned}
M^l_t=Softmax(Q^l_t\cdot {K^l_t}^T), \quad
\hat{\phi}^l_t=M^l_t\cdot V_t^l
\end{aligned}
\end{equation}
where $Q^l_t,K^l_t,V_t^l$ are the projection of $\phi^l_t$. Then the cross-attention module will generate the final feature conditioned on text embedding as the input for the next layer. 

\subsubsection{Content Injection}
\ 
\newline
Given the content image $I^c$, we perform the DDIM inversion to obtain the noised latents $\{z^{c*}_{t}\}^{T}_{t=1}$. In the reverse process of the content branch, each layer of the denoising U-Net produces the residual feature $(\Delta \phi^c)^{l}_t$ and the input of self-attention $(\phi^c)^l_t$. To inject content information, we replace the residual feature $\Delta \phi^{l'}_t$ in the target branch with $(\Delta \phi^c)^{l'}_t$ in the content branch, where $l'$ is the selected layer to apply residual replacement. The query and key elements of the self-attention will also be replaced at step $t$:
\begin{equation}
    (M^c)^l_t=Softmax((Q^c)^l_t\cdot {(K^c)^l_t}^T),  \quad
(\bar{\phi}^c)^l_t=(M^c)^l_t\cdot V_t^l
\end{equation}
where $(Q^c)^l_t, (K^c)^l_t$ is projected from $(\phi^c)^l_t$ in the content branch, while $V_t^l$ is projected from $\phi^l_t$ in the target branch. Note that the proposed content injection scheme involves two parts: (1) the residual replacement improves the preservation of high-frequency details, and (2) the attention replacement ensures consistency with the content image for the overall layout. Both are crucial for preserving the essential content structure (see Figure \ref{content injection}).
% With The proposed content injection scheme, the stylized image can preserve the essential content details. 

Step-by-step denoising has been observed with the phenomenon of error propagation \cite{li2023error}, which leads to poor reconstruction. To maintain the structure layout of $I^c$, the content branch takes the DDIM inversion latent $z^*_{t}$ as the input of the denoising U-Net at each step $t$. Simply, we formulate the reverse process in the content branch $\epsilon_\theta(z_t^{c*},t,\varnothing)$, where $\varnothing$ represents the null text. In this branch, U-Net is conditioned on the null text to mitigate the structure deviation caused by text-based noise prediction. Notice the input latent for the next step $t-1$ will be $z_{t-1}^{c*}$ obtained from DDIM inversion, instead of the U-Net's output $\hat{z}_{t-1}$ at the previous step $t$. 
 
\begin{figure*}[htbp]
\centering
\includegraphics[width=\linewidth]{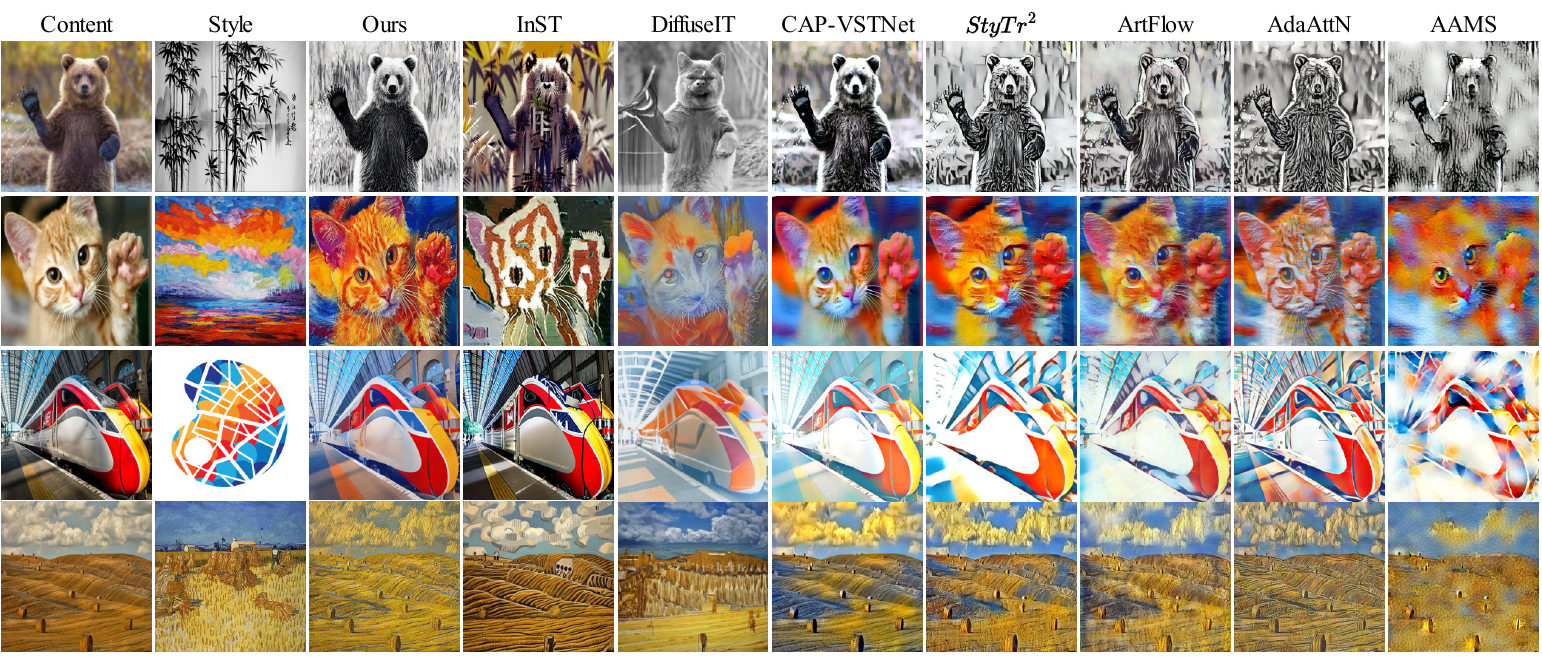}
\vspace{-20pt}
\caption{We compare our method with advanced style transfer methods. The content and style images are given in the first and second columns. Our method produces harmonious and high-quality stylized images, while others are less aesthetic due to the lack of balance between content and style representation.}
\label{qualitative}
\vspace{-1mm}
\end{figure*}

\subsubsection{Style Injection}
\
\newline
With the content injection module, we can prevent the problem of excessive styles in the generated images. However, the spatial content representation and the textual style representation are not perfectly compatible with each other to achieve a balance (see Figure \ref{style injection}). In this case, we introduce the style injection module with style spatial features to enhance the fusion. 

Similar to the content branch, we obtain the noised style image via DDIM inversion process as $\{z^{s*}_{t}\}_{t=1}^T$. Then we perform null-text reverse to extract the intermediate features $(\phi^s)^l_t$ of self-attention in the style branch, and inject the style spatial feature into the target branch at step $t$ to replace the key and value elements: 
\begin{equation}
    (M^s)^l_t=Softmax(Q^l_t\cdot {(K^s)^l_t}^T),  \quad
(\bar{\phi}^s)^l_t=(M^s)^l_t\cdot (V^s)^l_t
\end{equation}
where $(K^s)^l_t, (V^s)^l_t$ is projected from $(\phi^s)^l_t$ in the style branch, while $Q^l_t$ is projected from $\phi^l_t$ in the target branch. Recall that we use the DDIM inversion latent as the input to the per-step denoising in the style branch, denoted $\epsilon_\theta(z_t^{s*},t,\varnothing)$. The denoising at the next step starts with the inversion latent $z_{t-1}^{s*}$ rather than $\hat{z}_{t-1}^{s}$.

Notice that we replace different self-attention elements during content and style injection (see the right of Figure \ref{network}). For style transfer, content requires structure preservation, while style shows artistic visual effects. The attention map of self-attention has been proven to be aligned with an image’s semantic layout, i.e., content structure \cite{hertz2022prompt, tumanyan2023plug}. Therefore, we replace the query and key elements with the counterparts of the content image in the content branch, thus aligning the spatial layout of the generated image to that of the content image. This replacement choice of the content injection module has been verified in Appendix ~\ref{content_injection_elements}. To inject style, however, replacing key and value elements is more suitable to retain the content and encourage artistic detail in the generated image. 

\subsection{Separate Injection in Diffusion}
\label{unified}
Different steps of pre-trained diffusion models during the reverse process have been observed with varied functions \cite{meng2021sdedit, xu2023inversion, zhang2023prospect}, To produce a stylized image with balanced content and style, we are motivated to leverage the step-by-step nature of the diffusion model and separate the content and style injections in the target branch. 

Empirically, the early phase of the reverse process for the pre-trained diffusion model influences the overall structure and spatial layout, while the later phase focuses on more detailed visual effects. Therefore, we are inspired to perform the content injection in the early phase and the style injection in the later phase. Specifically, we introduce a hyper-parameter $\alpha$ and compute the deciding point $t^\alpha=\alpha\cdot T$ to decide the injection proportion. The reverse process in the target branch with separated injections is formulated as:
\begin{equation}
\label{reverse}
 \hat{z}_{t-1}=\begin{cases}\epsilon_\theta(\hat{z}_t,t,v^{st};(\phi^c)^l_t,(\Delta \phi^c)^{l'}_t)\quad &t>t^\alpha \\
 \epsilon_\theta(\hat{z}_t,t,v^{st};(\phi^s)^l_t)\quad &t \leq t^\alpha \end{cases}
\end{equation}
where step $t\in (0,T]$, and the initial latent for the target branch is the noised content image as $\hat{z}_T=z^{c*}_T$, which can ensure the generation with better preservation of the content structure. Based on the separate scheme, the noised codes required in the content branch are $\{z^{c*}_t\}_{t=t^\alpha+1}^T$, while in the style branch are $\{z^{s*}_t\}_{t=1}^{t^\alpha}$.

% When extending to the editing task, we use the text embedding $v^{edit}$ instead of $v^{st}$, and inject the spatial representations separately to produce $\hat{z}^{edit}_t$ using Equation (\ref{reverse}). Finally, the generated images are obtained by $I^{st}=\mathcal{D}(\hat{z}_{0})$ and $I^{edit}=\mathcal{D}(\hat{z}^{edit}_{0})$, respectively.

\section{Experiment}
\subsection{Experimental Setting}
Our method is built upon the official pre-trained BLIP-Diffusion \cite{li2024blip}. During sampling, we perform DDIM \cite{ho2020denoising} with 50 denoising steps, and set classifier-free guidance \cite{ho2022classifier} scale to 7.5. We set $\alpha=0.2$, $l$ as the decoder layers 4-11 to apply content and style injections, and $l'$ as the decoder layers 3-8 to inject residual features in the content branch. More implementation details are in the Appendix.

\subsection{Comparison with State-of-the-art Methods}
We conduct qualitative and quantitative comparisons with advanced diffusion-based methods as InST \cite{zhang2023inversion} and DiffuseIT \cite{kwon2022diffusion}, transformer-based methods as StyTr$^2$\cite{deng2022stytr2}, and methods in other architectures as CAP-VSTNet \cite{wen2023cap}, ArtFlow \cite{an2021artflow}, AdaAttN \cite{liu2021adaattn}, and AAMS \cite{yao2019attention}. 

\subsubsection{Qualitative Comparison}
\
\newline
Figure \ref{qualitative} is the visual comparison. The results of InST suggest that the learned textual embedding involves redundant structural information of the style image. Another diffusion-based method, DiffuseIT, catastrophically reconstructs the content information. For example, the panda in the first row has been changed to other creatures in the stylized images. The remaining methods all present a balance problem. CAP-VSTNet and ArtFlow produce photorealistic rather than artistic images.
% which are blended with colors of both content and style images, which are opposite to the goal of style transfer. 
% ArtFlow produces images with poor edge smoothness because of the numerical overflow. 
AdaAttN achieves content preservation, but its injection of artistic styles is inadequate and inaccurate. Nor do AAMS and StyTr$^2$ succeed in coexisting style and content representations harmoniously, as their generated images show strong artifacts or mosaic-like patterns.
% Especially when given style images with bright colors, their generated images (the last three rows) apparently fail to fuse the target styles. 
In general, the above methods fail to balance the content and style information.

In contrast, our proposed DiffuseST achieves a more balanced and high-quality stylization with harmonious and aesthetically pleasing results. For instance, the generated image in the first row can be observed with clear and distinct content of the bear, while successfully capturing the abstract style of ink painting. Another typical example is the cat in the second row, showing DiffuseST can blend the oil painting style with the content harmoniously without artifacts. Injecting style features as key and value elements into the self-attention module of the target branch, DiffuseST blends styles and contents based on their correlation, which ensures that the two kinds of features are compatible during generation.
% Recall that we inject style features as key and value elements of the self-attention module in the target branch. In this case, the fused styles are based on the correlation with the spatial features of the generated image, which means that the style and content features are compatible during the injection. Therefore, our method achieves harmonious and aesthetic synthesis.

\subsubsection{Quantitative Comparison}
\
\newline
% The numerical evaluation of style transfer is flexible As the ground truth images are not available.
We adopt the perceptual metric $LPIPS$ \cite{zhang2018unreasonable} to measure the consistency between the content image and the generated image. To assess style injection, we utilize the image-image CLIP similarity $CLIPscore$ \cite{radford2021learning} that calculates the cosine distance between the embedding of the target style image and that of the generated image. We also adopt the content loss $\mathcal{L}_c$ and the style loss $\mathcal{L}_s$.

As listed in Table \ref{quantitative}, we randomly select 40 content images and 20 style images to generate 800 images for a fair evaluation. As to content preservation, CAP-VSTNet achieves the best $LPIPS$ score and our method is the second-best. However, we emphasize that CAP-VSTNet fails to inject sufficient artistic styles, as it obtains the worst score in $CLIPscore$. Conversely, our method achieves the best score in terms of $CLIPscore$ for style injection. Other methods, such as DiffuseIT and StyTr$^2$, perform well in $CLIPscore$, however, show relatively high scores in $LPIPS$, indicating an inability to strike the balance. 
% Note that DiffuseST is comparable to learning-based methods in content loss. Although the style loss of DiffuseST is relatively higher than non-diffusion-based methods, we emphasize these methods are trained to minimize these losses. 
In addition, our method outperforms the two diffusion-based methods InST and DiffuseIT in all metrics. 
% This quantitative comparison verifies the stability of the proposed DiffuseST at the balance between content and style. 

We also conduct a user study on our method against four popular style transfer approaches: diffusion-based methods InST and DiffuseIT, Transformer-based method StyTr$^2$, and CNN-based method CAP-VSTNet. The setup details of the user study are in the Appendix. CAPVST-Net obtains more votes than ours in terms of content preservation. However, our method outperforms all other methods in style injection and overall quality. Note that although the diffusion-based method InST is second to our method on automatic metrics, it is far inferior to ours on human evaluation.

In Table \ref{runtime}, we compare three diffusion-based methods for the required time to generate one image. The results show DiffuseST with great advantage in runtime across other diffusion-based methods, nearly $\times 10$ faster than DiffuseIT, $\times 240$ faster than INST.

\begin{table}[tp]
  \caption{Quantitative comparison. $L$ for LPIPS and $C$ for CLIPscore. The best result is in bold and the second is underlined.}
  \vspace{-3mm}
  \label{quantitative}
  \setlength{\tabcolsep}{0.3mm}{
  \begin{tabular}{lcccccccc}
    \toprule
    & Ours & InST & DiffuseIT & CAP & StyTr$^2$ & ArtFlow & AdaAttN & AAMS \\
    \midrule
    $L \downarrow$ & \underline{0.26} & 0.28 & 0.45 & \textbf{0.22} & 0.42 & 0.39 & 0.36 & 0.51\\
    $C \uparrow$ & \textbf{0.60} & \underline{0.59} & 0.58 & 0.50 & 0.53 & 0.50 & 0.56 & 0.57\\
    $\mathcal{L}_c \downarrow$ & \underline{1.49} & 2.22 & 3.08 & \textbf{0.86} & 1.98 & 1.73 & 2.18 & 2.40\\
    $\mathcal{L}_s \downarrow$ & 8.69 & 12.53 & 9.86 & 5.31 & \textbf{2.02} &3.72 & \underline{2.74} & 4.14\\
  \bottomrule
\end{tabular}}
\end{table}

\setlength{\belowcaptionskip}{0pt}
\begin{table}[tp]
  \setlength{\tabcolsep}{0pt}
  \caption{User study. The listed scores are the percentage of votes from 50 participants that the compared methods are preferred to ours, in terms of content preservation, style injection, and overall image quality.}
  \vspace{-3mm}
  \label{User study}
  \setlength{\tabcolsep}{2mm}{
  \begin{tabular}{ccccc}
    \toprule
    & InST & DiffuseIT & CAP-VSTNet & StyTr$^2$ \\
    \midrule
    $Conetnt$ & 28$\%$ & 11$\%$ & 51$\%$ & 42$\%$ \\
    $Style $ & 37$\%$ & 21$\%$ & 39$\%$ & 35$\%$ \\
    $Overall$ & 31$\%$ & 5$\%$ & 35$\%$ & 39$\%$ \\
  \bottomrule
\end{tabular}}
\vspace{-2mm}
\end{table}
\setlength{\belowcaptionskip}{0pt}
\begin{table}[tp]
  \setlength{\tabcolsep}{0pt}
  \caption{Runtime comparison of diffusion-based methods.}
  \vspace{-3mm}
  \label{runtime}
  \setlength{\tabcolsep}{2mm}{
  \begin{tabular}{cccc}
    \toprule
    & Ours & InST & DiffuseIT \\
    \midrule
    Runtime (s) & $\sim$5 & $\sim$1200 & $\sim$48\\
  \bottomrule
\end{tabular}}
\vspace{-2mm}
\end{table}

\subsection{Ablation Study}
In this section, we investigate two main schemes in our proposed method, which are (1) the parameter $\alpha$ to separate injections, and (2) the textual and spatial representations.
\subsubsection{Analysis of Hyper-parameter $\alpha$}
\begin{figure}[tp]
\centering
\includegraphics[width=\linewidth]{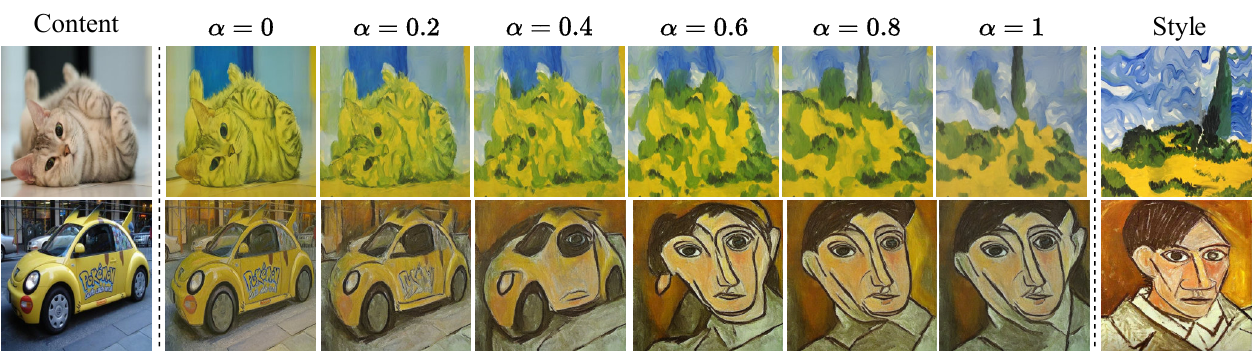}
\vspace{-6mm}
\caption{Ablation study on $\alpha$ to control the injection proportion of content and style. Larger $\alpha$ determines more steps in the target branch for style injection.}
\label{alpha}
\end{figure}

\
\newline
We introduce a parameter $\alpha$ to decide different inject steps for content or style spatial features. The deciding point is computed as $t^\alpha=\alpha\cdot T$, where content injection module will be applied for the early denoising stage as steps from $T$ to $t^\alpha+1$, while style injection from steps $t^\alpha$ to $1$. Figure \ref{alpha} shows the effect of $\alpha$ for the stylization balance. 
% The generation is conditioned on the textual embedding $v^{st}$ for all values of $\alpha$. 
Note that all images are conditioned on the same textual embeddings $v^{st}$.
When set $\alpha$ to $0$, the whole reverse process is applied with content injection without any style injection. In this case, the generated images in column 2 to a large extent are similar to the content images. It also suggests that the textual style information of embedding $v^{st}$ by the BLIP-2 encoder is insufficient to strike the balance. Similarly, the generated images of $\alpha=1$ in column 7 with full style injection could be severely biased to the style images while the content structures are unrecognizable. Empirically, we select $\alpha=0.2$ for a stable and robust stylization quality. Yet, users can choose other $\alpha$ values to satisfy their personal preferences for varying the strength of the injected artistic styles.

\subsubsection{Analysis of Textual and Spatial Representations}
\
\newline
To verify the effectiveness of combining textual and spatial representations, we conduct the following experiments to investigate their effects on the generated image.

\begin{figure}[tp]
\centering
\includegraphics[width=\linewidth]{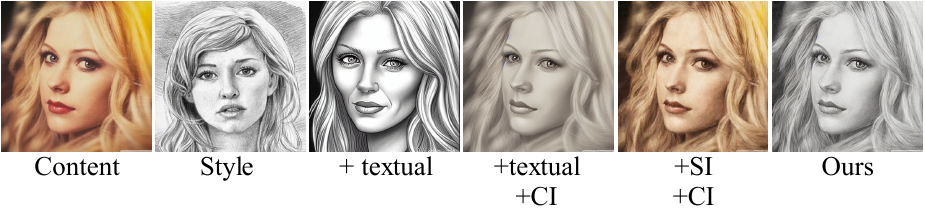}
\vspace{-6mm}
\caption{Ablation study of textual and spatial representations. Together with textual condition, content injection (CI), and style injection (SI), DiffuseST (the last column) achieves harmonious and balanced stylization.}
\label{style injection}
\end{figure}

Figure \ref{style injection} studies the effectiveness of the textual embedding by BLIP-2 \cite{li2024blip} encoder and the two spatial feature injections. Solely conditioned on the extracted text embedding $v^{st}$ of the style image, without any injection of spatial features, the image in column 3 shows incomplete content structure, and the density of the sketch style is not aligned with the target one. Conditioned on $v^{st}$ with content injection solely, the image in column 4 shows that the content information can dominate the stylization, which verifies the necessity to inject style spatial representation. Using two injection modules while removing the learned textual embedding, the image in column 5 presents a faded yellow color rather than the desired sketch style, demonstrating the effectiveness of our proposed textual representation. With all modules, the image in column 6 (Ours) can present an accurate sketch feeling and intact facial features.

\begin{figure}[tp]
\centering
\includegraphics[width=\linewidth]{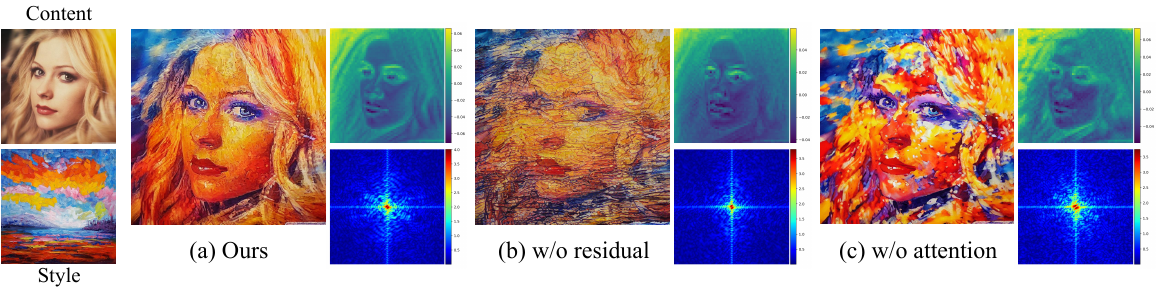}
\vspace{-6mm}
\caption{Ablation study of content injection module with residual and attention feature injections. We visualize the output self-attention features of the decoder layer $l=9$ at step $t=420$ and analysis in Fourier space.}
\label{content injection}
\end{figure}

Figure \ref{content injection} studies two components in the content injection module: (1) the residual feature injection, and (2) the attention feature injection. Image (b) without the residual feature injection lacks high-frequency components in the Fourier space, resulting in an unrecognized face contour. This observation concurs with a discovery that residual features play an important role in learning critical spatial content \cite{zhang2018residual, liu2020residual, liu2024cdformer}. When removing the attention feature injection from the content branch, image (c) contains high-frequency information, however, neglects some details such as hair, which is crucial for content preservation. In contrast, image (a) with the injection of both residual and attention features achieves satisfying content consistency and style alignment. 

In the Appendix, we provide additional ablation experiments, including the injection order (see Appendix ~\ref{injection Order}), the hyper-parameter $l,l'$ to apply injection (see Appendix ~\ref{injection_layer}), and the replacement choice of the content inject module (see Appendix ~\ref{content_injection_elements}).

\subsection{Extensions}
\begin{figure}[tp]
\centering
\includegraphics[width=\linewidth]{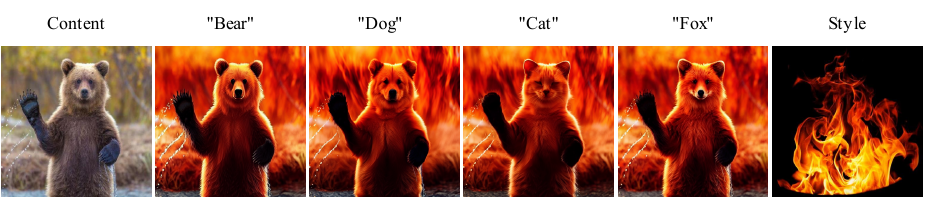}
\vspace{-6mm}
\caption{Extend DiffuseST to image editing by modifying the content text $\mathcal{P}^c$ to $\mathcal{P}^{edit}$ (e.g., "dog"). }
\label{text_edit}
\end{figure}

In Figure \ref{text_edit}, we show the text-based image editing results of DiffuseST. We manually state a content text $\mathcal{P}^{edit}$ as other animals, then replace the content embedding $v^c$ with $\hat{v}^c=\psi(\mathcal{P}^{edit})$. For instance, in column 3 we use $\mathcal{P}^{edit}=``dog"$ to modify the the content structure of the stylized image. The generated images show changed animals, while maintaining the overall layout and the target artistic style. This suggests that DiffuseST has inherited pre-trained text-to-image diffusion models' ability to the potential of editing. Appendix ~\ref{extensions} provides image-to-image translation results via DiffuseST.

\section{Conclusion}
In this work, we propose a novel training-free approach for style transfer to achieve balanced and controllable stylization. Our method combines the textual and spatial representations of the content and style images to accurately capture the artistic styles and preserve the essential content details. For textual representation, we employ the multi-modal BLIP-2 encoder to predict the text-aligned embedding of the style image. For spatial representation, we utilize the DDIM inversion technique to extract intermediate features from the content and style branches. Then we propose content and style injection modules that modulate the spatial features in the target branch, and apply them in the reserve process separately to balance the content and style. We conduct various comparison and ablation experiments to demonstrate the efficiency and effectiveness of the proposed method. Our method achieves high-quality synthesis with a balance between content preservation and style injection. Moreover, DiffuseST retains the controllability and editability of pre-trained diffusion models, with the potential to satisfy user-specific preferences for aesthetic appearance and unique styles.

\section{Acknowledgements}
This work was supported in part by the Natural Science Foundation of China (No. 62272227), and the Postgraduate Research \& Practice Innovation Program of NUAA (No. xcxjh20231604).


\bibliographystyle{ACM-Reference-Format}
\bibliography{reference}

\appendix
\section{Preliminary}
\label{sec:Preliminary}
\subsection{Text-to-Image Diffusion Models}
Generative diffusion models \cite{rombach2022high, song2020denoising} produce high-quality images from Gaussian noise by iterative denoising. To train a diffusion model, the forward process gradually adds Gaussian noise:
\begin{equation}
\begin{aligned}
z_t = \sqrt{\alpha_t}z_0+\sqrt{1-\alpha_t}\epsilon, \epsilon \sim \mathcal{N}(0,1)
\end{aligned}
\end{equation}
where $z_0$ is the image sampled from the training data,  $\epsilon$ is the added Gaussian noise, and $\alpha_t:= \prod_{i=1}^{t}(1-\beta_i)$ is the hyper-parameter determined by the schedule $\beta_0,...,\beta_T\in (0,1)$.

The reverse process of the diffusion model $\epsilon_\theta$ progressively predict the noise $\epsilon$, where the objective is defined as:
\begin{equation}
\begin{aligned}
E _{z_0,\epsilon\sim\mathcal N(0,1),t}[\Vert\epsilon-\epsilon_\theta(z_t,t) \Vert_2^2]
\end{aligned}
\end{equation}

In this work, we are particularly interested in pre-trained Latent Diffusion Models (LDMs) conditioned on text prompts. Instead of the image $x_0$ sampled from the real data distribution, the source latent $z_0$ is encoded as $z_0=\mathcal{E}(x_0)$. The image can be generated by the decoder as $x=\mathcal{D}(\hat{z}_0)$, where $\hat{z}_0$ is the denoised latent by $\epsilon_\theta$. The text prompt $\mathcal{P}$ can be input at step $t$ as:
\begin{equation}
    \hat{z}_{t-1}=\epsilon_\theta(\hat{z}_t,t,v)
\end{equation}
where $v=\psi(\mathcal{P})$ is the conditional text embedding encoded by the pre-trained CLIP \cite{radford2021learning} text encoder $\psi$. 

DDIM \cite{song2020denoising} has been proposed for faster sampling by making the reverse process deterministic. Assuming that the ODE process can be reversed in several small steps, DDIM inversion \cite{song2020denoising, dhariwal2021diffusion} is further introduced to manipulate non-generated images by performing DDIM sampling in reverse order:
\begin{equation}
\begin{aligned}
\label{inversion}
z^*_{t+1} = \sqrt{\frac{\alpha_{t+1}}{\alpha_t}}z^*_t+(\sqrt{\frac{1}{\alpha_{t+1}}-1}-\sqrt{\frac{1}{\alpha_t}-1})\cdot\epsilon_\theta(z^*_t,t,v)
\end{aligned}
\end{equation}

\subsection{BLIP-Diffusion}
Vanilla text-to-image diffusion models enable the generation of high-quality images conditioned on text prompts, but cannot take images as input.
BLIP-Diffusion \cite{li2024blip} is a state-of-the-art model for the subject-driven editing task. It adopts BLIP-2 \cite{li2023blip} encoder to produce text-aligned visual features. Given the target image, it transforms visual representations to textual conditions by a feed-forward layer. In this work, we adopt the multimodal encoder in BLIP-Diffusion to project style images into text embeddings. The encoded features contain rich generic information and can express the semantic characteristics of style images. For simplicity, we denote the feature extraction by BLIP-2 encoder as:
\begin{equation}
    \bar{v}=\mathcal{F}(I,T)
\end{equation}
where $I$ is the input image, $T$ is the input text, and $\bar{v}$ is the text-aligned visual features of the image $I$.
\section{Algorithm}
The general algorithm of our method for the style transfer task is detailed in Algorithm \ref{algo:st}.

\begin{algorithm}
  \caption{Style Transfer}
  \label{algo:st}
  \textbf{Input: }{Pre-trained diffusion model $\epsilon_\theta$, text encoder $\psi$, BLIP-2 encoder $\mathcal{F}$, content image $I^c$, style image $I^s$, $\alpha$.}\\
  $v^{c}\leftarrow \psi(""), \bar{v}^{s}\leftarrow \mathcal{F}(I^s, "")$\;
  $v^{st}\leftarrow Concat(v^{c}, \bar{v}^{s})$\;
  $t^\alpha \leftarrow \alpha\cdot T$\;
  $z^c_0 \leftarrow \mathcal{E}(I^c)$, $\{z^{c*}_t\}_{t=t^\alpha+1}^T \leftarrow DDIM\_Inversion(z^c_0, \varnothing)$\;
  $z^s_0 \leftarrow \mathcal{E}(I^s)$, $\{z^{s*}_t\}_{t=1}^{t^\alpha} \leftarrow DDIM\_Inversion(z^s_0, \varnothing)$\;
  $\hat{z}_T \leftarrow z^{c*}_T$\;
  \For{$t\leftarrow T,...,1$}{
    \eIf{$t>t^\alpha$}{
      $(\phi^c)_t^l, (\Delta\phi^c)^{l'}_t \leftarrow \epsilon_\theta(z^{c*}_t,t,\varnothing)$\;
      $\hat{z}_{t-1}\leftarrow \epsilon_\theta(\hat{z}_t,t,v^{st};(\phi^c)_t^l, (\Delta\phi^c)^{l'}_t)$\;
    }{
      $(\phi^s)_t^l \leftarrow \epsilon_\theta(z^{s*}_t,t,\varnothing)$\;
      $\hat{z}_{t-1}\leftarrow \epsilon_\theta(\hat{z}_t,t,v^{st};(\phi^s)_t^l$\;
    }
  }
  \textbf{Output: }{$I^{st} \leftarrow \mathcal{D}(\hat{z}_{0})$.}
\end{algorithm}

\section{Implementation Details}
To conduct the comparison and ablation experiments, we adopt the images that appeared in \cite{johnson2016perceptual,wen2023cap,liu2021adaattn,an2021artflow}. All experiments are conducted on RTX 3090 in a single GPU. 
\begin{figure*}
\centering
\includegraphics[width=0.9\linewidth]{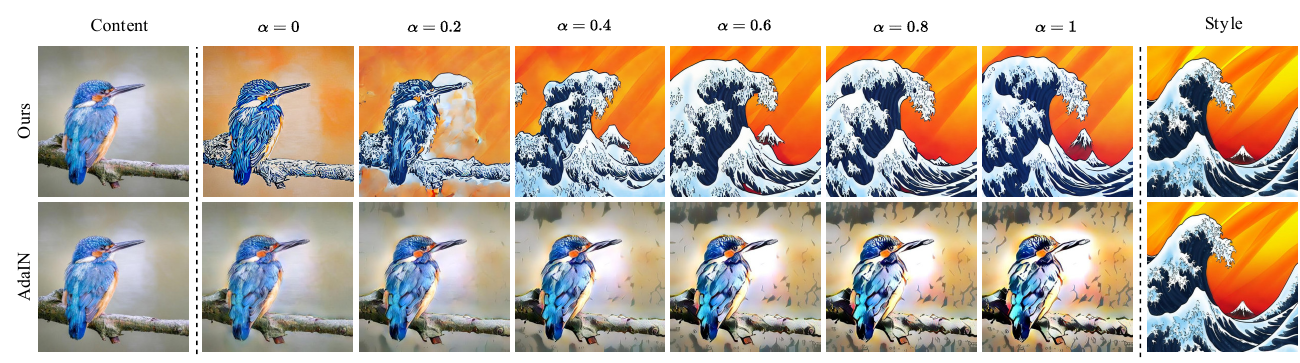}
\caption{Compare the controllability of stylization between our method and AdaIN. The stylized images of AdaIN with varied hyper-parameter $\alpha$ only show slight changes in artistic styles, which suggests that it modulates the basic content features and forcibly injects style information, rather than sensible fusion. In contrast, DiffuseST harnesses different feature representations and separates the two injections, controlling the strength of fusion towards a balanced stylized image in a true sense.}
\label{alpha_contrast}
\end{figure*}

\begin{figure*}
\centering
\includegraphics[width=0.9\linewidth]{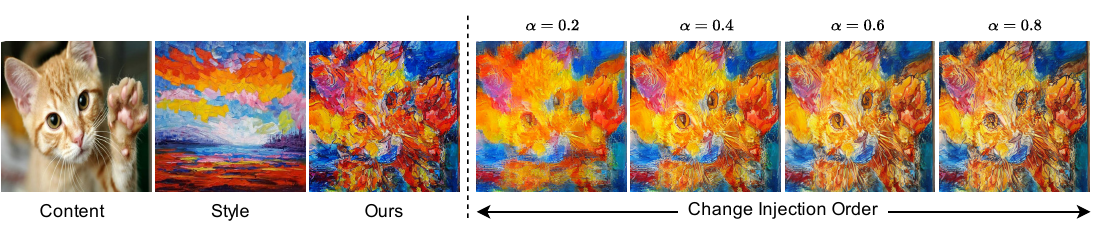}
\caption{We study the order for injection. The results are generated by applying the style injection from step $T$ to $t^\alpha+1$, and then the content injection from step $t^\alpha$ to $1$. The results indicate that injecting style features before the content injection is not suitable for style transfer. For example, the image in column 4 ($\alpha=0.2$) is applied with style injection in the early stage with more steps as 80 percent of the whole time interval, but it leads to the content structure of the cat being unrecognizable. The generated image in column 7 ($\alpha=0.8$) can not match the expected style of strong oil painting.}
\label{injection_order}
\end{figure*}

\section{Controllability Comparison}
The introduction of parameters to achieve controllable generation is not new to the style transfer task. For example, AdaIN \cite{huang2017arbitrary} tackles the content-style trade-off with the following equation:
\begin{equation}
    T(c,s,\alpha)= g((1-\alpha)f(c)+\alpha \mathrm{AdaIN}(f(c),f(s)))
\end{equation}
where $f$ is a pre-trained VGG-19 \cite{simonyan2014very}, $\mathrm{AdaIN}$ to produce the fused feature maps, the parameter $\alpha$ for style interpolation, and the generative network $g$ to output the stylized image $T$.

However, we do emphasize that the controllability of DiffuseST is distinguished from the above methods. As shown in Figure $\ref{alpha_contrast}$, the controllability of AdaIN is limited. The largest $\alpha$, which should produce the most stylized image, but AdaIN's result does not match the stylized image, either in structure or in artistic information. It demonstrates that AdaIN is constantly injecting styles into the original content image, rather than fusing content and style representations. In contrast, our method separates the injection of content and style information modules utilizing the time interval of diffusion models. In this case, the degree of stylization via DiffuseST can be adjusted using the parameter $\alpha$ that calculates the deciding point. The results of our method are obviously more consistent with the given style information as $\alpha$ increases. This comparison verifies the effectiveness and accuracy of the injection modules and the separation scheme we have designed.

\section{Ablation study}

\subsection{Injection Order}
\label{injection Order}
One of our key motivations is that the step-by-step nature of the pre-trained diffusion model can facilitate stylization to achieve the balance between content and style. We are inspired to separate the injection, applying the content injection in the early stage of the reverse process while the style injection in the later stage. To verify this choice of injection order, we test the exchanged order in Figure \ref{injection_order}. In this setting, given the deciding step $t^\alpha=\alpha\cdot T$, we perform the style injection from step $T$ to $t^\alpha+1$ while the content injection from step $t^\alpha$ to $1$. The results demonstrate that injecting style representation before content representation can not capture the desired artistic style accurately, nor preserve the content structure.

\subsection{Injection Layer Selection}
\label{injection_layer}
The layer $l$ to apply the attention feature injection is an important parameter in our method to achieve balanced stylization. In this section, we provide detailed ablation experiments to verify our selection of the layer for injection. 

\begin{figure}[t]
\centering
\includegraphics[width=\linewidth]{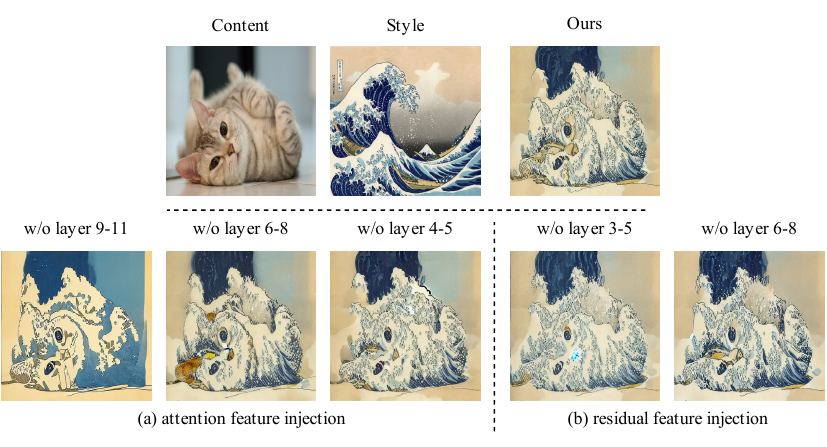}
\vspace{-6mm}
\caption{Ablation study for the decoder layer to inject content spatial representation. Feature size is $16\times16$ for layers 3-5, $32\times32$ for layers 6-8, and $64\times64$ for layers 9-11. (a) to inject attention features, all layers affect the generated image for content details. (b) to inject residual features, lower layers are more crucial for high-quality synthesis. }
\label{content_injection}
\vspace{-3mm}
\end{figure}

Both the encoder and decoder layers in the U-Net involve the residual, self-attention, and cross-attention features. Previous works \cite{tumanyan2023plug} have demonstrated that the features in the decoder layer are more critical for content preservation. Therefore, we follow them and apply the content injection module in the decoder layer. We then study the effect of different decoder layers in Figure \ref{content_injection}. We divide the decoder layers according to the resolution: layers 4-5 in resolution $16\times16$, layers 6-8 in resolution $32\times32$, and layers 9-11 in resolution $64\times64$. We find in practice that injecting attention features at the decoder layers 4-11 and residual features at the decoder layers 3-8 can produce high-quality stylized images.

\begin{figure}[t]
\centering
\includegraphics[width=\linewidth]{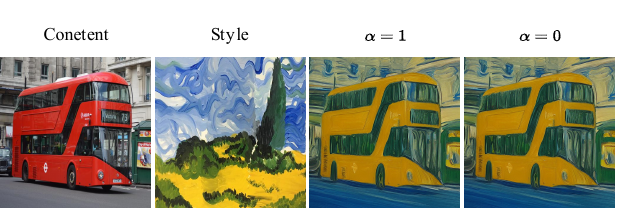}
\vspace{-6mm}
\caption{Inject style spatial features in the encoder layers. We vary the parameter $\alpha$ to amplify the strength of style spatial injection, i.e., $\alpha=1$ with full style injection. However, the generated image changes slightly compared to the result of $\alpha=0$, which indicates that the encoder layers are not suitable for style injection.}
\label{style_injection1}
\vspace{-2mm}
\end{figure}

\begin{figure}[htbp]
\centering
\includegraphics[width=\linewidth]{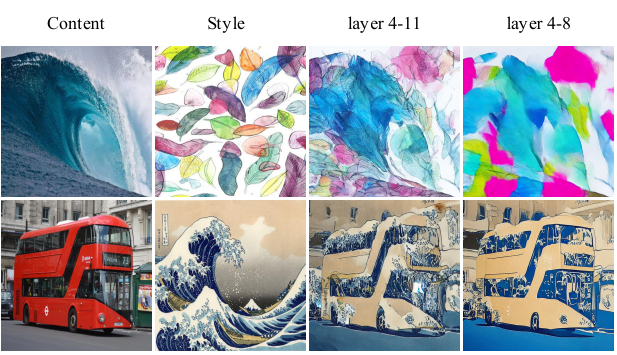}
\vspace{-6mm}
\caption{We study different decoder layers to apply style injection. The colors of the images in the last column without layers 9-11 (i.e., the style features in resolution $64\times64$) apparently deviate from the style image.}
\label{style_injection2}
\end{figure}

While style spatial representations may function differently from the content representation, we also try the encoder layer to apply the style injection module. As shown in Figure \ref{style_injection1}, the results suggest that the replaced features in the encoder layers only have slight effects. Even when we set $\alpha$ to 1 to maximize the influence strength of style injection, the generated results fail to change as significantly as we expect. Then we consider the decoder layer to inject style representation, and study the effect of different layer choices in Figure \ref{style_injection2}. We find that the $64\times64$ features ensure accurate injection of the style image's colors. Empirically, we set the injection layers for style representations as the decoder layers 4-11.


\subsection{Content Injection Elements}
\label{content_injection_elements}

\begin{figure}
\centering
\includegraphics[width=0.9\linewidth]{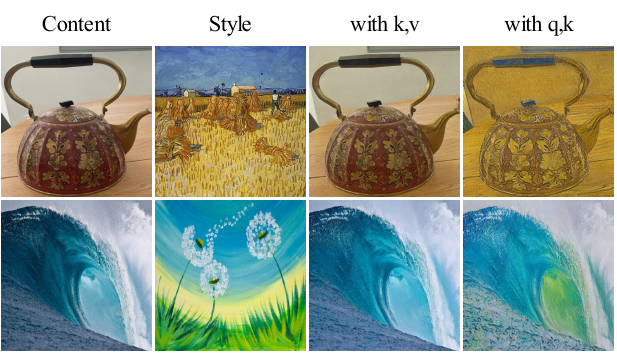}
\caption{Ablation study of the choose with q,k and k,v.}
\label{qualitative_content_module}
\end{figure}

We have conducted an ablation study of replacing which elements on the content injection module. In Table \ref{quantitative_content_module}, we quantitatively compare replacing the key and value elements (k,v) or replacing query and key elements (q,k) in the self-attention module, the content/style loss of replacing k,v is 0.62/11.38, while replacing q,k resulted in 1.49/8.46. This comparison suggests that replacing k,v fails to inject adequate style details (extremely high style loss but low content loss). Though replacing q,k shows a higher content loss, it is the balanced stylization that we are concerned about. In Figure \ref{qualitative_content_module}, we show the qualitative results for two settings. It is obvious that replacing k,v injects insufficient style into the content image (e.g., the pot is still realistic in row 1 column 3). 

\begin{table}[h]
  \caption{Quantitative comparison. $\mathcal{L}_c$ for the content loss and $\mathcal{L}_s$ for the style loss.}
  \vspace{-3mm}
  \label{quantitative_content_module}
  \setlength{\tabcolsep}{2mm}{
  \begin{tabular}{lccc}
    \toprule
    & inject k,v & & inject q,k (ours) \\
    \midrule
    $\mathcal{L}_c \downarrow$ & 0.62 & & 1.49\\
    $\mathcal{L}_s \downarrow$ & 11.38 & & 8.46\\
  \bottomrule
\end{tabular}}
\end{table}

\section{Extensions}
\label{extensions}
\begin{figure}[htbp]
\centering
\includegraphics[width=\linewidth]{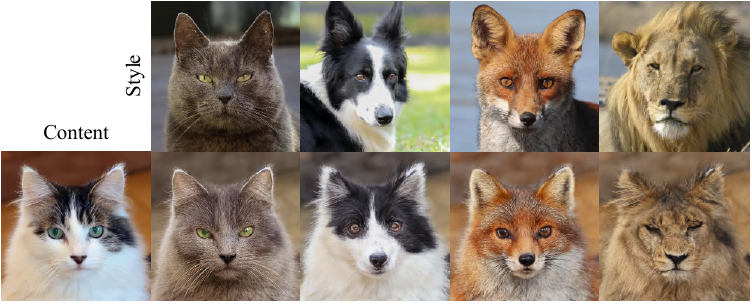}
\caption{Image-to-image translation results via our method.}
\label{Domain adaptation}
\end{figure}

Our approach can also be applied to image-to-image translation, as shown in Figure \ref{Domain adaptation}. We input the content and style images as both animal images from the dataset AFHQ \cite{choi2020stargan} to obtain the adapted results. For instance, we select a cat as the content image and a dog as the reference image to generate the image in column 3, which can show the feeling of the dog and is in line with the cat in spatial layout.

\section{User Study}
We asked 50 participants to rank all compared methods with the following questions: 1) Rank the following images according to better content preservation, 2) Rank the following images according to better style injection, and 3) Rank the following images according to the synthesis quality. We randomly selected 10 images from the generated 800 images by each method. Finally, we count the votes that existing methods are preferred to ours.

\section{Limitation}
\begin{figure}[t]
\centering
\includegraphics[width=\linewidth]{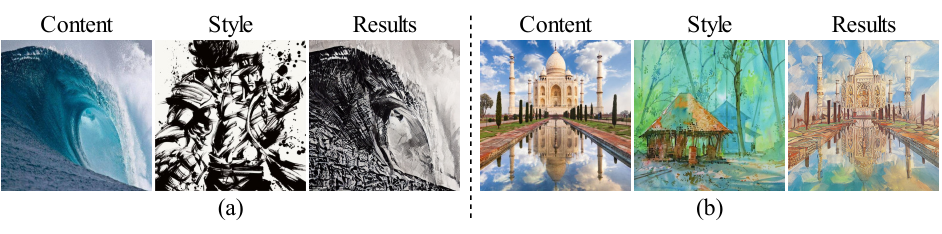}
\caption{Limitations. (a) DiffuseST fails to achieve balanced stylization given the style image with a strong artistic feeling. (b) The colors are not consistent between the generated and the style images due to attention calculation.}
\label{drawback_1}
\end{figure}

Although DiffuseST achieves high-quality style transfer in most cases, we acknowledge that it is subject to the following limitations: (1) given strong artistic styles, our method may favor the content picture and show inharmonious stylization; (2) the stylized image may deviate from the color distribution of the given style image, which may be caused by moderate activations during attention computation. Figure \ref{drawback_1} shows three failure cases about the above limitations. In the first row, the ink style does not harmonize with the content of the wave in the stylized image. In the second row, the primary color of the style image, i.e., green color, is not evident in the stylized image.
% some content information such as the village at the foot of the hill disappears due to the strong style of injection. In the first two lines, the colour distribution of the generated result deviates a little from the provided style image.} 

\begin{figure}[htbp]
\centering
\includegraphics[width=\linewidth]{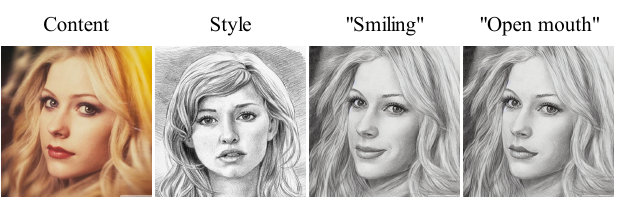}
\caption{Limitation of our method. The edit prompts are limited and may fail to generate the desired edited image when largely changing the layout of the content image.}
\label{drawback}
\end{figure}

Another discovery is that extended text-based image editing can only work on certain attributes or contents. As shown in Figure \ref{drawback}, we extend our method for the text-based image editing task and modify the emotion of the content image as a woman. In column 3, we input the editing prompt $\mathcal{P}^{edit}=$"Smiling", and obtain an edited image as expected. However, when adding the attribute "Open mouth" which requires more changes in spatial layout, our method fails to produce the desired image as in column 4. We hypothesize that the injection manner of the content spatial features is too tough to change the original layout away from the original content image.

\section{More Results}
\begin{figure*}[hp]
\centering
\includegraphics[width=\linewidth]{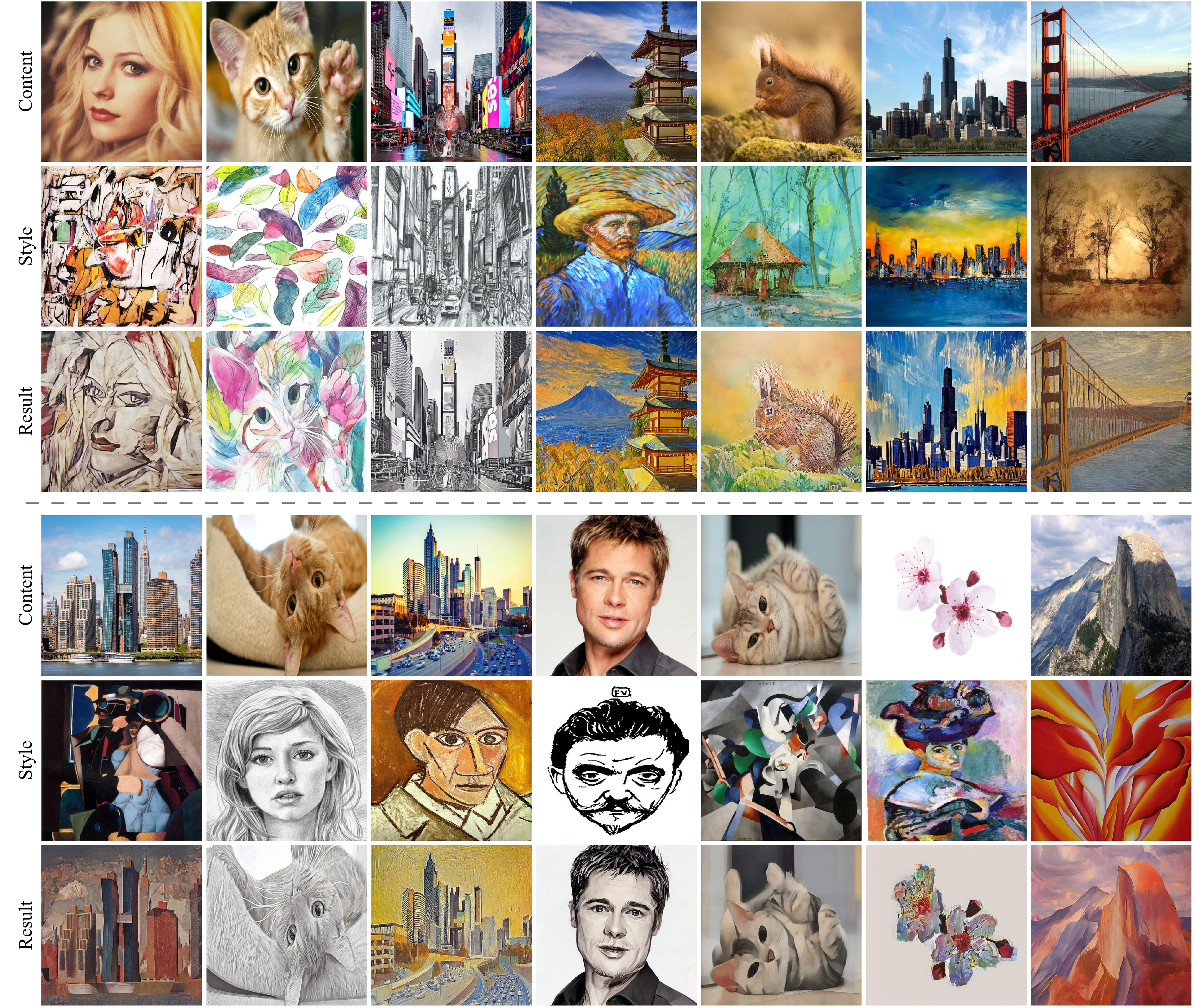}
\caption{Additional style transfer results by our method.}
\label{more_results_1}
\end{figure*}

\begin{figure*}[hp]
\centering
\includegraphics[width=0.85\linewidth]{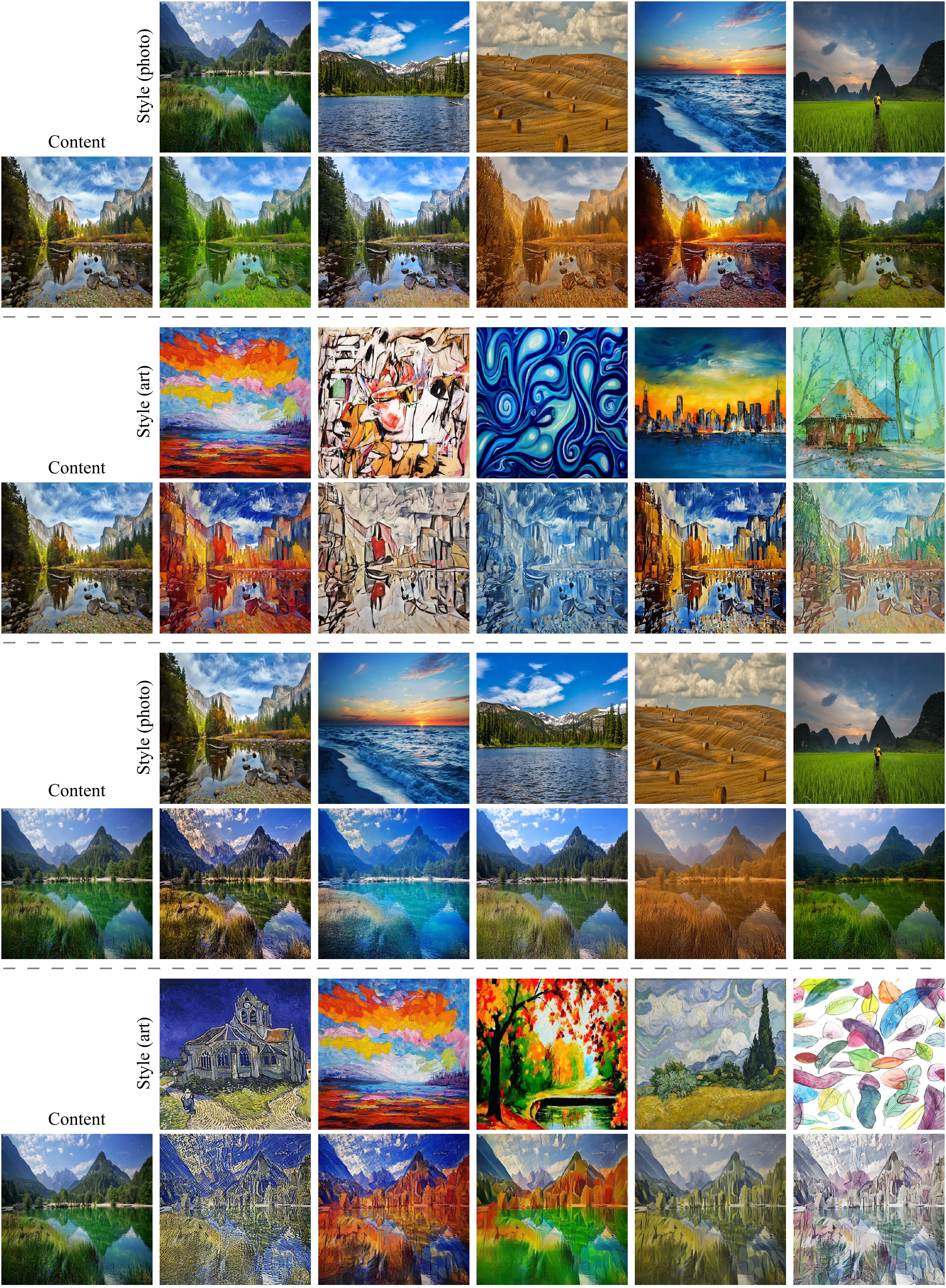}
\caption{Additional results by our method with different style types such as photorealistic or art style.}
\label{more_results_2}
\end{figure*}

We provide additional qualitative results as follows:

(1) Figure \ref{more_results_1} and \ref{more_results_2} are more generated results of our method for style transfer in different content images (e.g., photorealistic landscape images) and diverse style images (e.g., abstract art).

\begin{figure*}[hp]
\centering
\includegraphics[width=\linewidth]{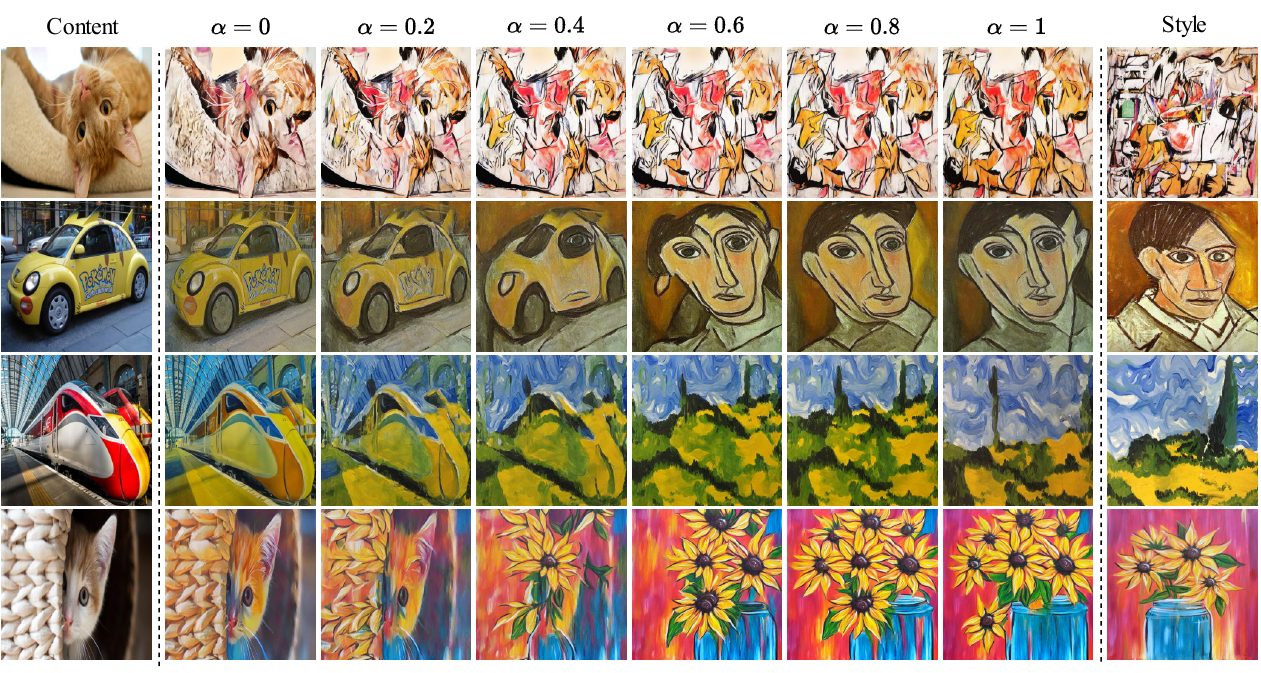}
\caption{We provide more examples for the ablation study of hyper-parameter $\alpha$.}
\label{more_results_alpha}
\end{figure*}

(2) Figure \ref{more_results_alpha} is the additional result for varying the hyper-parameter $\alpha$ to control the strength of style.

\begin{figure*}[hp]
\centering
\includegraphics[width=\linewidth]{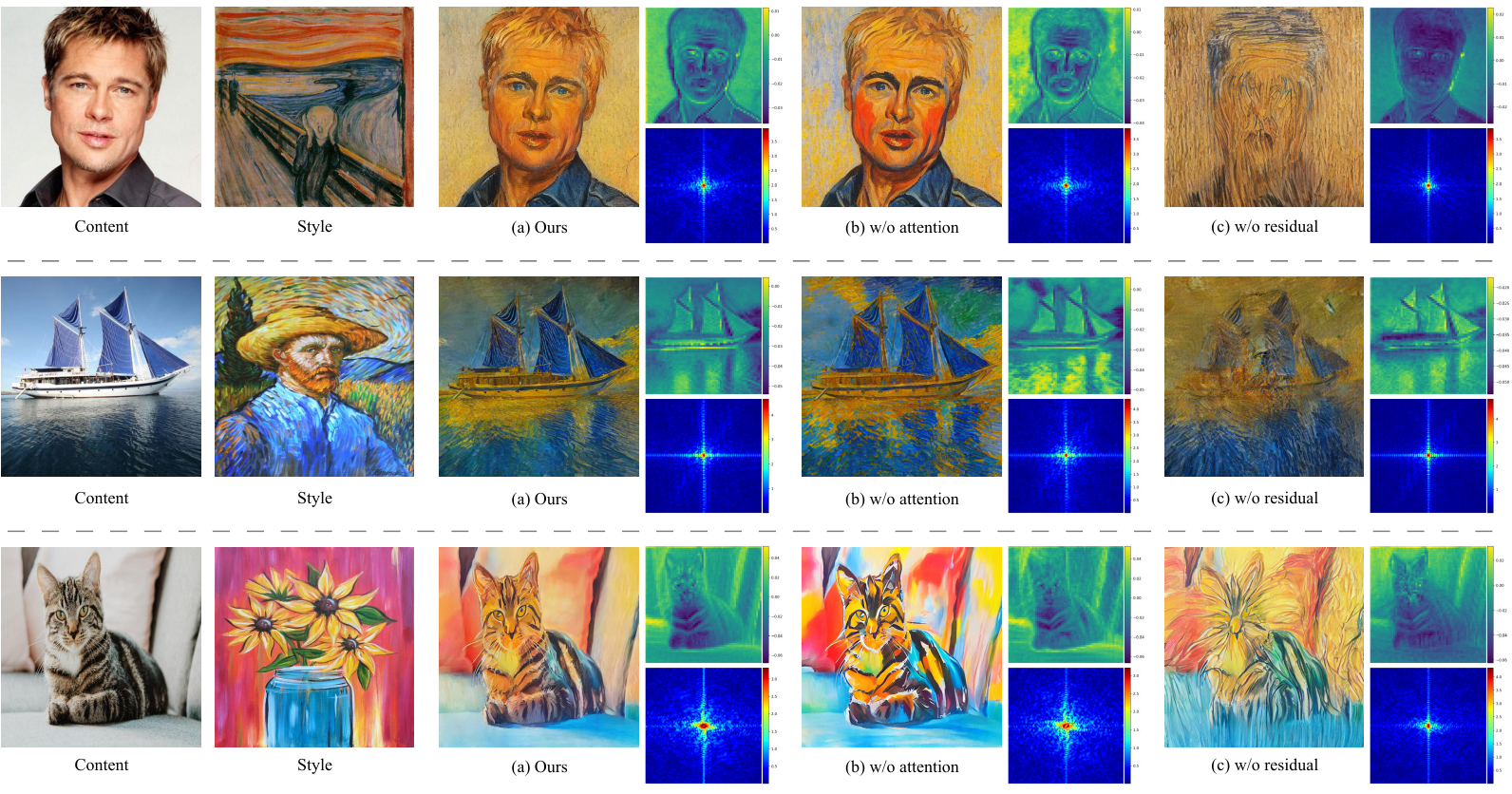}
\caption{We provide more examples for the visualization of feature maps and analysis in Fourier space.}
\label{appendix_feature_vis}
\end{figure*}

(3) Figure \ref{appendix_feature_vis} is the additional result for feature maps visualization about the content injection ablation experiment.

\begin{figure*}[hp]
\includegraphics[width=\textwidth]{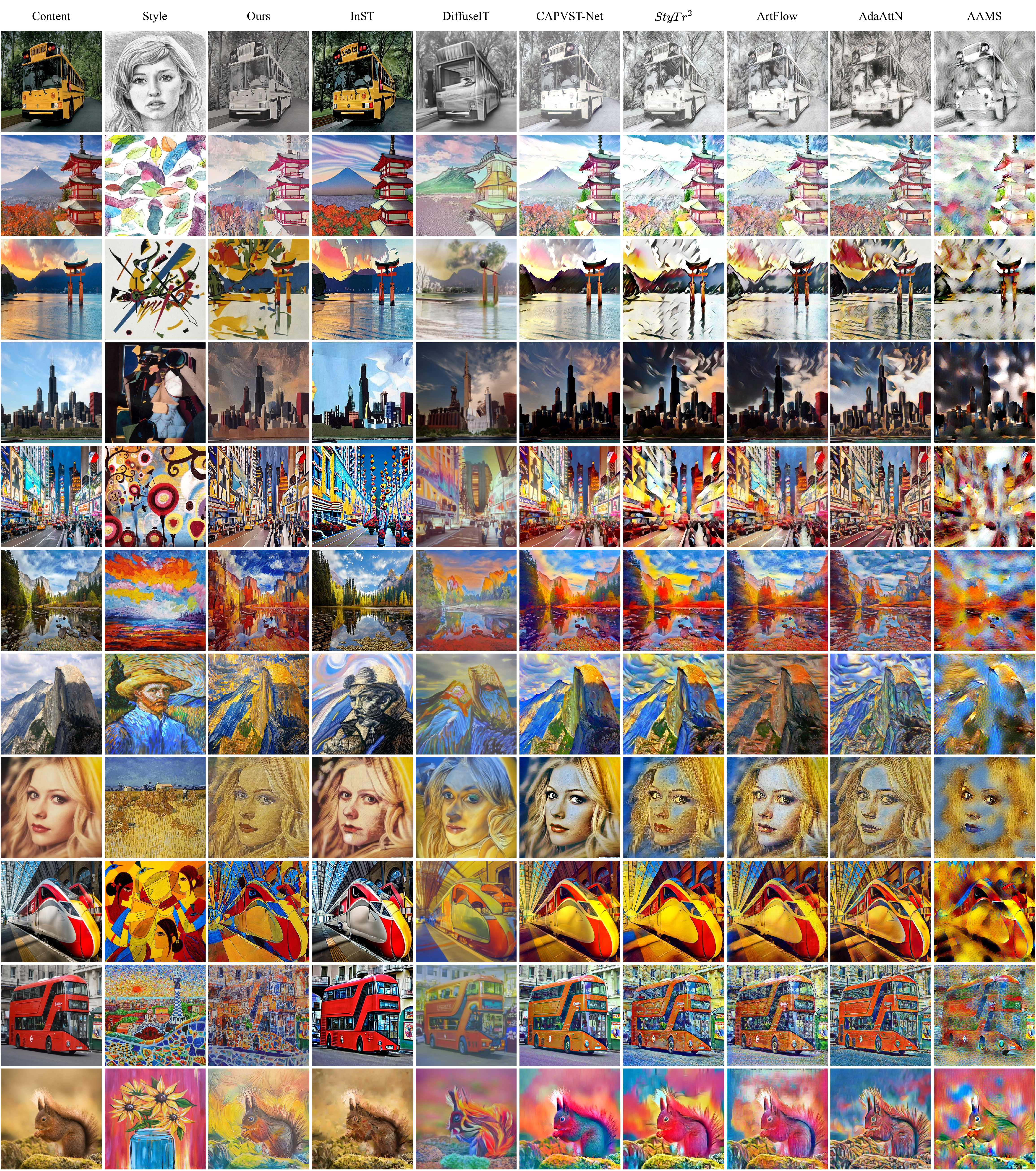}
\caption{Additional qualitative comparison results.}
\label{more_results_comparison}
\end{figure*}

(4) Figure \ref{more_results_comparison} is additional qualitative comparison result.

\end{document}